\documentclass[review]{elsarticle}

\usepackage{lineno,hyperref}
\usepackage{amsmath}
\usepackage{amsfonts}
\modulolinenumbers[5]

\journal{Journal of \LaTeX\ Templates}

\begin{document}
\nolinenumbers

\begin{frontmatter}

\title{Memory as a \textit{Mass-based} Graph:\\ 
	 Towards a Conceptual Framework for the Simulation Model of Human Memory in AI
}

\author[mymainaddress]{M. Mollakazemiha} \cortext[mycorrespondingauthor]{Corresponding author}
\ead{mollakazemiha.mahdi@gmail.com}
\author[mysecondaryaddress]{H. Fatzade}

\address[mymainaddress]{Faculty of Mathematical Sciences, Shahid Beheshti University, Tehran 1983969411, Iran.}

\address[mysecondaryaddress]{Department of Philosophy, Faculty of Humanities, University of Zanjan, Zanjan 4537138791, Iran.}

\begin{abstract}

There are two approaches for simulating memory as well as learning in artificial intelligence; the functionalistic approach and the cognitive approach. The necessary condition to put the second approach into account is to provide a model of brain activity that contains a quite good congruence with observational facts such as mistakes and forgotten experiences. Given that human memory has a solid core that includes the components of our identity, our family and our hometown, the major and determinative events of our lives, and the countless repeated and accepted facts of our culture, the more we go to the peripheral spots the data becomes flimsier and more easily exposed to oblivion. It was essential to propose a model in which the topographical differences are quite distinguishable. In our proposed model, we have translated this topographical situation into quantities, which are attributed to the nodes. The result is an edge-weighted graph with \textit{mass-based} values on the nodes which demonstrates the importance of each atomic proposition, as a truth, for an intelligent being. Furthermore, it dynamically develops and modifies, and in successive phases, it changes the mass of the nodes and weight of the edges depending on gathered inputs from the environment.

\end{abstract}


\end{frontmatter}

\nolinenumbers

\section{Introduction}

\noindent
The early postulated models of human memory date back to 1976 when Tulving supposed for finding something in memory there is a serial search process in your brain. When you are seeking something for recall from the box of storage, you go to that particular storage box and search for each content at a time until you pick your desired one (Tulving, 1976). By the development of numerical methods to simulation of human memory, models like MINERVA 2 (Hintzman, 1984), CHARM (Eich, 1982), TODAM (Murdock, 1993), the Matrix model (Humphreys et al., 1989), and SAM (Raaijmakers \& Shiffrin, 1981), found themselves strongly related to the global matching models (Hintzman, 1984; Humphreys et al., 1989; Kahana et al., 2005; Kelly et al., 2017; Clark \& Gronlund, 1996).
While any proposed model has its specific structure and terminology, resemblances in the mathematics behind these models provide evidence of a general agreement on this topic. In order to reach a unified theory of human memory, Matthew A. Kelly and Robert L. West, suggested a theoretical framework for identifying each proposed memory model within six key decisions namely; “(1) choice of knowledge representation scheme, (2) choice of data structure, (3) choice of associative architecture, (4) choice of learning rule, (5) choice of time-variant process, and (6) choice of response decision criteria” (Kelly \& West, 2017). The representation scheme of human memory model starts with the LISP (List Processing) in which cognitive architecture ACT-R (Anderson \& Lebiere, 2014) and SAM (Search of Associative Memory) models are represented as “storage and retrieval of discrete symbols” (Clark., 2001). Symbolic models are represented as “expressions of symbols and the manipulation of those symbols” (Clark., 2001; Kelly et al., 2017), which they are inspired by linguistics and logic (Locke \& Phemister, 2008). In the symbolic model, “the atomic units of the model are concepts (\textit{e.g., dog, glowing, noun, or anger}) and complex units arise from characterizing the relations between those concepts.” (Kelly \& West, 2017). By the time of 1980 Neurally-inspired connectionist models were suggested in which the atomic units are “artificial neurons or simple neuron-like nodes, whereas a concept will have complex implementation as a pattern of activation across those units” (McClelland et al., 2010). In vector-based models of memory; vectors serve as a \textit{to be remembered item} (Gayler, 2004; Plate, 1995). Vector Symbolic Architectures (VSAs) was the promoted version of the previous model; computational associative memories were used by cognitive psychologists in order to model both behavioral and neurological aspects of human memory. In that model, with some mathematical operations, symbols or vectors (which are representing symbolic information) could be modified or combined within the so-called sub-symbolic process (Kelly et al., 2013). Such models developed, and evolved at the same time, from semantic to syntactic types in which they infer the meaning of words from "how the words co-occur in a corpus" (Kelly, Reitter, et al., 2017). In a comprehensive sense, there are two approaches in the simulation of memory and learning in artificial intelligence; in the functionalistic approach, without understanding the events inside the mind, a proposed model attempts to intake inputs and deliver outputs that correspond to the outputs delivered by the mind to those inputs (Clark., 2001). It is also important, due to the complex relations between inputs and outputs, to propose an algebraic system that should be capable of anticipating upcoming events; The operators of this algebraic system are a functionalistic representation of unknown events within the mind. In the simpler version of this viewpoint, one can create a connection between inputs and outputs only by the mean of a computational method without building an algebraic system (Putnam, 1988). In the model arising from this viewpoint, it does not matter those events within the mind have fundamental differences from the corresponding events in artificial intelligence. The notable point in these models is the simulation of the \textbf{apparent function} and the \textbf{ultimate behavior} of the mind. In fact, the mind is considered as a black box which its internal logic can be understood by an accurate and detailed vision throughout the relationships between inputs and outputs (Rouse \& Morris, 1986), and in some respects, basically, the meaning of "understanding" is nothing but perceiving these observational relationships. Any other system with the same function has the equivalent internal logic. In this view, which is consistent with behaviorism in the realm of psychology, the mind is seen as a function whose identity is determined by the relationships between inputs and outputs. Subsequently, proposed algorithms establish mathematical relations between data which can predict the system’s behavior either deterministically or stochastically. The result is that this system, despite its correspondence with the activities of the mind, is fundamentally different from it; does not make mistakes or forget. This is the perfectness of simulated systems that makes us skeptical of them and gives rise to the idea that this is exactly why we can think, and we have the will and the emotions. which, unlike artificial intelligence, we are not perfect. In the functionalist approach, we seem to transcend the limitations of the mind and eliminate its shortcomings, but in fact, we replace the mind with something different. Which, although it has some of its functions, is something quite different from the mind, and in the strict sense of the word, we cannot talk about simulation. These inevitable constraints of this approach persuaded scientists to pursue a more complicated cognitive approach.
In cognitive approach, it is not merely sufficient to simulate the relationships between inputs and outputs, but it is important to decode the contents of the black box of the mind using methods like introspection and empathy and by resorting to psychological theories like the hermeneutic method and simulate its internal activity in artificial intelligent. For instance, the idea of a self-learning network in which an artificial intelligence gathers inputs directly or in relationship with other intelligence, modifies its prior information over time, and thus acquires a history, has emerged from this approach.
The cognitive approach has recently become more prominent because of the advanced progress of cognitive science which we have seen in the last two or three decades.

\section{Method}

In this article, with the same cognitive approach, we try to put this memory phenomenon into perspective and suggest a model that can be the basis for simulating memory in artificial intelligence. This model, as expected, will be a dynamic model that continuously improves in successive phases. Due to the limitation of memory space and the necessity of differentiating between matters of high importance, less importance, and non-importance, some information will be stored permanently in long-term memory and some others forget or disappear sooner. Our proposed dynamic model seeks to simulate this human phenomenon. If artificial intelligence is supposed to make the same civilization, the imperfection of mathematics would be inevitable. Otherwise, we are inescapable to include emotions, feelings, complexities, and elegancies of humanity as initial values or the same function of them or other conditions into artificial intelligence. Thus, we missed simulating the most important aspect of intelligence called "volition and freedom". Our goal is to propose a very simple \textit{étude} of non-algorithms that could one day be the basis of creating imperfect and erroneous artificial intelligence. A forgetful artificial intelligence that does not progress linearly, and correctness or incorrectness of its decisions depends on its lifestyle and the events surrounding it, often experiences and grows, but sometimes it also regresses and loses some of the right components of its decision-making. And that is exactly why you can love or hate it because these give her feeling and freedom. To this end, we first introduce a novel mathematical concept that will be the basis of our proposed model.

\section{Mass-based Graph}

Since in prevailing theories, declarative memory stores propositionally. Thus, we considered the bricks of our model as atomic propositions in which the connections between them form a semantic network that represents our memory structure. Note, though, the "image" is perceived by Gestalt principles, but every perceived image, for understanding in the form of meaning, decomposes into propositions that are understood linearly and diachronic. In Wittgenstein's words, we consider each atomic proposition to be equivalent to a state of affairs. On the other hand, the “meaningfulness unit” which was considered as a “concept” in Greek thought, is recognized as a “proposition” in modern philosophy (Quine, 1997). Therefore, we started our simulation from the propositional level. To illustrate this structure, we use the concept of graph which is conventional. But since our proposed model is dynamic, the regular graphs are not suitable for our model, for this purpose, we introduce a new graph called "mass-based graph".

\begin{figure}[ht]
	\includegraphics[height=5cm]{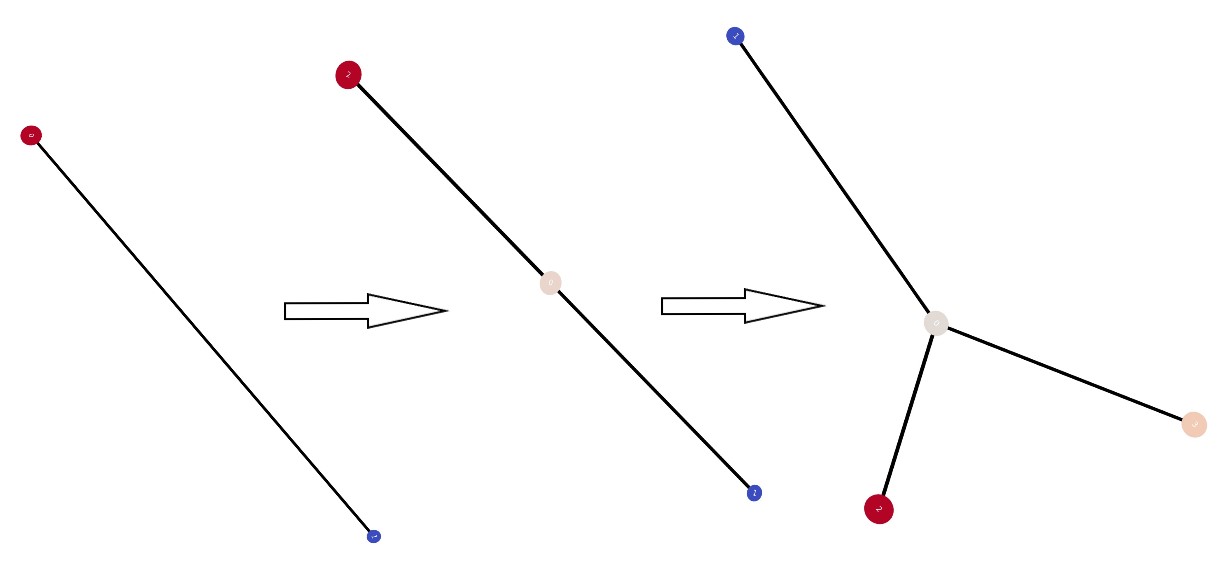}
	\caption{Evolution of the Dynamic Mass-based Graph from phase 1 (left hand side) to phase 3. By adding each node and edge, the size(mass) and thickness(weight) of nodes and edges change respectively. Initial mass-based values and initial weights of each node and edge have been chosen randomly from the standard normal distribution.}
\end{figure}

In this graph, we assign an initial mass-based value to each node which are representing the degree of importance for an intelligent being. In the following, the edges make the connections between these nodes, in which, they already contain initial weight values regarding their \textit{environmental} conditions.  By adding any new edge, the importance and therefore the mass of the nodes on both sides increases, thus, this leads to an increase in the weight of all connected edges to these two nodes. Accordingly, the graph will have a special mode in each level (phase) in which, by adding each new weighted edge, there will be a great change in the next level. It should be noted that only one edge can be drawn between the two nodes, and we show the importance and \textit{thickness} of that edge with its weight (Figure 1). With these explanations, we introduce the technical viewpoint of this mass-based graph.

The graph in its primary state, formed by nodes and edges with initial mass and weight values. In our model, the initial mass-based value of each node (atomic proposition) corresponds to its importance to an intelligent being. We can classify these initial mass-based values and their importance into three main categories for the human being. The most important category of these propositions relates to our survival. For evolutionary reasons, all the facts that relate to our biological survival have special places in memory, and we rarely forget them (like sexual activity or intimacy (Lindau et al., 2018)). Another category of these propositions is related to our identity and they are becoming important to our recognition. This category of propositions is a psychological type of the first category of propositions, and our survival and mental health are based on them. While the truth about nutrition belongs to the first category, the event that led to our humiliation belongs to the second category of propositions and they remain as drastic and precise as possible in our memory. In addition to these two categories, a third category should be considered as a kind of cultural specialty. While this category of propositions is not capable of being delivered to the natural originality of two previous categories, inevitably, the consequence will be a kind of fetishism. In this case, what was supposed to be a tool in line with the goals of the previous two categories, becomes an ultimate and independent goal and is fixed in memory for its own sake. This variation in the initial mass-based values provides many possible choices for shaping artificial intelligence ahead of us. In the next section, we will see that the initial inputs in the adjacency matrix are responsible for this diversification.

Regarding the initial weight of the edges, it seems that the situation of their occurrence is a determining factor. In fact, the initial weights of the edges, which illustrate the causal relationships or correlations or co-occurrence of propositions, are situation dependent; a situation that can be emotional or cultural. Indeed, the repetition and attention that in the classical view should be considered as a main factor in weighting the edges, in our model, treated as a second-class of importance which is stemmed from the mass of the nodes, i.e., the importance of the relevant facts.

Now, after the initial establishment of the graph, it is time to turn it into a dynamic graph with an introduction to the graph rule:
\begin{quote}
	\textit{\textbf{Graph Rule:} By adding each new edge with a certain initial weight value (which we have assigned earlier), this edge possesses an increase in its weight proportional to the average mass of its nodes. The mass of the nodes also increases proportional to the initial weight, and this increase in mass leads to an increase in the weight of all other edges associated with these two nodes.}
\end{quote}

\begin{figure}[ht]
	\includegraphics[height=5cm]{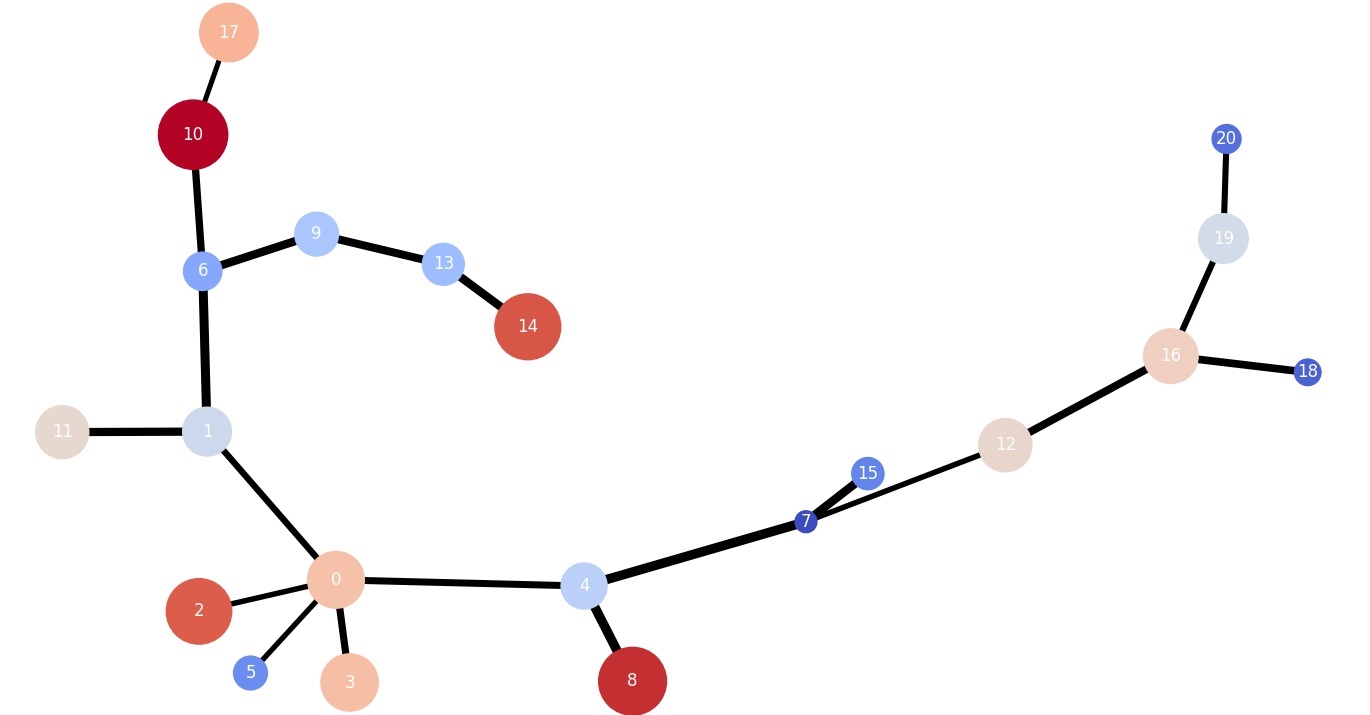}
	\caption{Mass-based Dynamic Graph after 22 phase for randomly chosen initial mass-based and wight values. Connections were made randomly to demonstrate the dynamicity of the graph.}
\end{figure}

Accordingly, by adding each new weighted edge, we will encounter a great change in the state of the graph system, which this new state will be the basis for entering the next phase. As a result, a graph with the arrival of any new edge in each phase takes on a dynamic shape (Figure 2). At the same time, it is possible to extend the graph by defining a new node with a certain initial mass-based value. But this expansion in the nodes is not related to the dynamism of the graph and this is merely a simple expansion that can be applied to any other static graph.


\section{Introducing the model} 
Consider our graph formed by n nodes from $M_1, M_2, \cdots,M_n$. We consider the values of $M^0_{1}, M^0_{2}, \cdots, M^0_{ n}$ as initial mass-based values. Because of the use of the logarithmic function, we suppose these values to be greater than 1. Then consider the zero-diagonal symmetric matrix of order $n$, called the matrix of the edge which elements represent the initial weight of the edge that corresponds to the related nodes. For the same reason, the values of the weight of the edges must be greater than $1$. In this way, the inputs of the graph will be as follows:
$$
\bordermatrix{~ & M_1^0 & M_2^0 & M_3^0 & \cdots & M_n^0 \cr
	M_1^0 & 0 &  W_{12}^0 & W_{13}^0 & \cdots & W_{1n}^0 \cr
	M_2^0 & W_{21}^0 & 0 & W_{23}^0 & \cdots & W_{2n}^0\cr
	M_3^0 & W_{31}^0 & W_{32}^0 & 0 & \cdots & W_{3n}^0\cr
	\vdots& \vdots & \vdots & \vdots & \ddots & \vdots \cr
	M_n^0 & W_{n1}^0 & W_{n2}^0 & W_{n3}^0 & \cdots & 0\cr}\,,
$$
Thus, this graph has $\frac{n(n+1)}{2}$ input number greater than $1$ that $n$ of the inputs are associated with initial mass-based values and $\frac{n(n-1)}{2}$ of the inputs are for initial weight values. These initial inputs are called phase zero of the graph.

Now we get output from these inputs, which will indicate the graph state in its first phase. In contrast to the initial condition of phase zero, in this phase, data become meaningful. In another word, both nodes regarding their geopolitical importance in this graph described as their masses and edges, find their appropriate weights attribute to their position in the graph as a consequence of determining real mass-based values of nodes. In this model, although the phases are discrete, they represent the variable of time. As mentioned above, it seems that the learning (storing) and recall (retrieve) of information in the human mind are also discrete and obtained in quantifiable packet level by level. The final state of the first phase is as follows:
\begin{eqnarray*}
M^1_{i}&=&M^0_{i}+\sum_{k=1}^{n}f(W^0_{ki})\,,\quad f(0)=0\,,\quad f(x)=\ln x\ + g(x;\mu ,\sigma )\,,\quad(x>1)\\
W^1_{ij}&=&W^0_{ij}+\ln(M^1_{i}+M^1_{j})\
\end{eqnarray*}
Where \( g(x;\mu ,\sigma ) \) is:
\begin{eqnarray*}
\frac{1}{{x}\pi \sigma [1 + (\frac{\ln({x}) - \mu  }{\sigma  })^{2}] }\;,\quad \forall {x}\in \mathbb{R} \mid {x}>{1}
\end{eqnarray*}
Which is Log-Cauchy probability density function with parameters \( \mu  \)  and \( \sigma  \) these two parameters can be arbitrary chosen or estimated.  One has to choose an appropriate defined domain for this function after properly estimating the initial input numbers. The selected interval should ultimately do not have a significant impact on the next phase or two.  

The reason that why we use the logarithmic function is that the function must be monotone increasing on its domain while its rate of growth decreases gradually such that the repetition effect cannot override the fundamental importance including biological, psychological, and fetishism. We also intuitively expect that in the beginning, the repetition effect to be more impressive and gradually decrease with increasing repetition. Neurological studies have also shown that when the stimulus is repeated, the BOLD response in fMRI imaging in the second demonstration is less than the first time, which is called the repetition suppression effect (Gazzaniga et al., 2014). Moreover, the decreasing intensity in the growth rate of the logarithmic function requires the necessity of assigning the initial values, for both weights and masses, in not large numbers, say from 2 to 100. In that case, summation of the logarithms of weights of the connected edges to a specific node would be a reasonable value in comparison with the mass of that particular node. Additionally, it is necessary for each of these \textit{intelligences} to have their own character, and not to have so-called \textit{mass production}. We add this characteristic, which can indicate a lack of causation in the realm of the mind, to what is known as the condition for the happening of free will, with the Cauchy distribution function. This function, which represents the seventh level of randomness in Mandelbrot's theory, is added to the formula with little effect, in order to be determinant only at the boundary conditions. We know that normal people in intermediate conditions, have causal and predictable and formulable behavior, and their differences and characteristics are often revealed in vague boundary conditions. It can be said that our character is drawn within our borders. The presence of this random function sometimes causes a small, insignificant (low mass) node to be deleted and forgotten in one step, and sometimes, the same node remains and And its mass increases in other phases, and it may even become an important (high mass) node in the future. 

\begin{figure}[ht]
	\includegraphics[height=6.5cm]{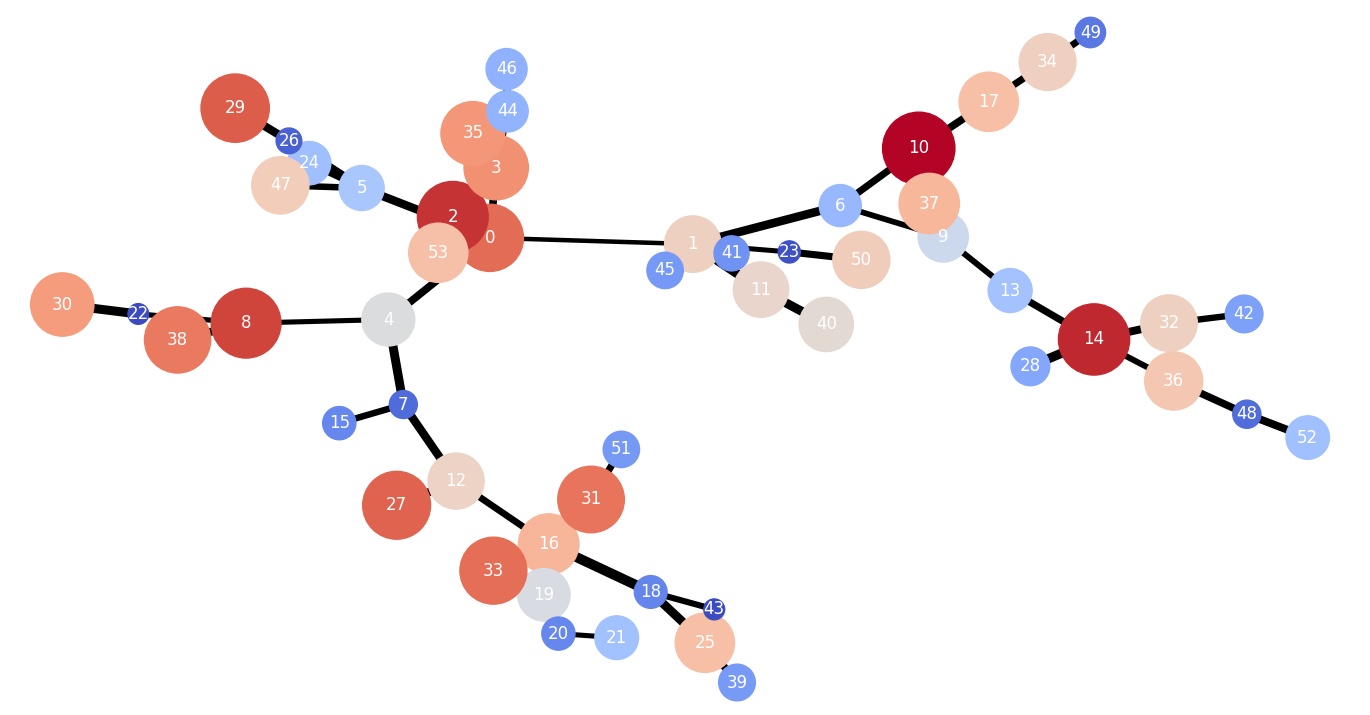}
	\caption{Mass-based Dynamic Graph after 55 phases of randomly      chosen initial mass-based and weight values. Notice how size    of the nodes (mass-based values) and thickness of the edges     (weight values) have changed.}
\end{figure}

Now we determine the state of this graph in phase $t~ (t > 1)$ (Figure 3). Agreeably, we define variable m as follows: $m = n$ , This variable will be used in phases where a new node is added instead of an edge.

For this purpose, we consider a number greater than 1 as the input, indicating the initial weight of new edge. This entry is called \textit{dynamic input}. Suppose this new edge connects the nodes $k$ and $l~ (1 \le k, l \le m)$. $\overline{W}^t_{kl}$ demonstrates the initial weight value of this edge which counts as its initial weight. Since this is a new edge, we have the following important condition:
$$
\forall s<t\,,\quad W^s_{kl}=0\,.
$$
The graph state in the $t$ phase is calculated as follows:
\begin{eqnarray*}
	\mathrm{Edge~ weight~ in~ phase~} t:\qquad W^t_{kl} &=& \overline{W}^t_{kl} + \ln \left(M^{t}_{k} + M^{t}_{l}\right)\\
	     \mathrm{Mass~ of~ nodes~ in~ phase~} t:\quad M^t_{k} &=& M^{(t-1)}_{k} + f(\overline{W}^{t}_{kl})\\
	     M^t_{l} &=& M^{(t-1)}_{l} + f(\overline{W}^{t}_{kl})\\
	     M^t_{i} &=& M^{(t-1)}_{i}\quad (i \neq k, l)
\end{eqnarray*}
The weight of the other edges associated with the nodes, due to the increase in the mass of the two nodes:
\begin{eqnarray*}
W^t_{kp} &=&\left\{
\begin{array}{ll}
W^{(t-1)}_{kp} + \ln \big(M^t_{k}-M^{(t-1)}_{k}\big)	&	;W^{(t-1)}_{kp}\neq 0\,\,,p \neq l\\
0 & ;W^{(t-1)}_{kp}=0
\end{array}\right. 
\\
W^t_{pl} &=&\left\{
\begin{array}{ll}
W^{(t-1)}_{pl} + \ln \big(M^t_{l}-M^{(t-1)}_{l}\big)	&	;W^{(t-1)}_{pl}\neq 0\,\,,p \neq k\\
0 & ;W^{(t-1)}_{pl}=0
\end{array}\right.
\end{eqnarray*}

\begin{figure}[ht]
	\includegraphics[height=6.2cm]{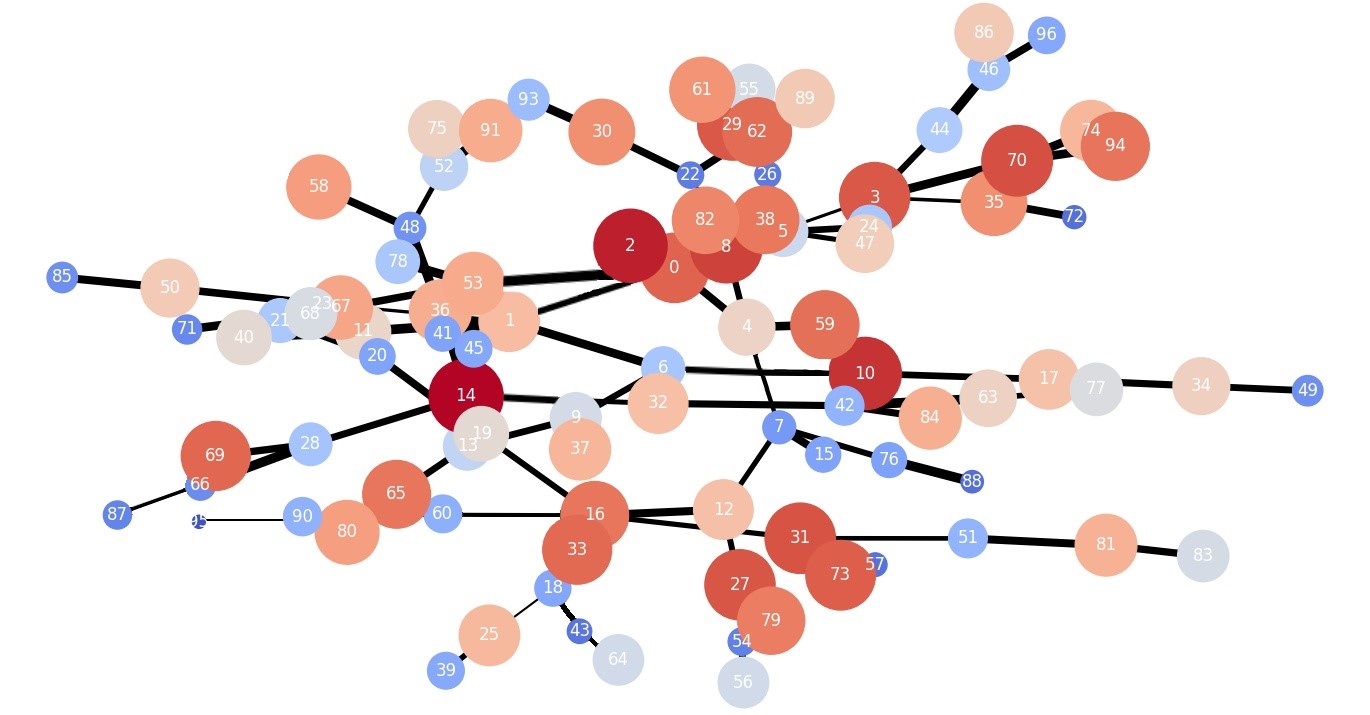}
	\caption{Mass-based Dynamic Graph after 100 phases. Primarily nodes such as node $1$, $2$, $3$, etc., mostly became dense and distinguishable. Some other marginal nodes, due to the existence of Log-Cauchy distribution, gained much more mass-based values and were remained as crucial nodes.}
\end{figure}

where $ 1 \le p \le m$. Thus, in each phase, the weight of the edges determines which connections are more important and vital and which will be gradually eliminated. Energy limits will determine the threshold of this elimination. The point is that a node which is isolated and no further edge runs out of it will also be deleted \textit{and involved players in the game will continue to play}. Note that in phase $t$ it is plausible that a new node with an initial mass-based value of $M_{t,m+1}$ adds into the graph. In this state, we won’t encounter any particular change but a static development of the graph and the creation of new possibilities of adding new edges. Notably, with the addition of each new node, we have: $m = m+1$ (Figure 4).

\section{Discussion and conclusion}
The brilliant of this model is its capability to account for being both an algorithm for supervised/unsupervised machine learning (such as clustering or its usage for NLP (Natural Language Processing) or an algorithmic tool for optimization in search engines. At the same time, with more stochastic ingredients, it opens a vast majority of possibilities of conducting future studies in the field of cognitive modelings, in particular, memory retrieval tasks. The result of such cognitive models could demonstrate the importance and the role of each proposition in the memory network or the drift rate (i.e. the difficulty of retrieval) of each specific proposition (node). 
Thus, the mass-based graph with its dynamicity for changing the shape of itself sounds promising for being a conceptual framework for future studies in related fields in the hope of advancing technology and human civilization.

\section*{Declaration of Competing Interest}
The authors declare no conflict of interest.

\section*{Acknowledgments}
This research did not receive any specific grant from funding agencies in the public, commercial, or
not-for-profit sectors.


\end{document}